\def\year{2022}\relax

\documentclass[letterpaper]{article} 
\usepackage{aaai22}  
\usepackage{times}  
\usepackage{helvet}  
\usepackage{courier}  
\usepackage[hyphens]{url}  
\usepackage{graphicx} 
\usepackage{natbib}  
\usepackage{caption} 
\DeclareCaptionStyle{ruled}{labelfont=normalfont,labelsep=colon,strut=off} 
\usepackage{microtype}
\usepackage{amsmath}
\usepackage{subfigure}
\usepackage{algorithm}
\usepackage{algpseudocode} 
\usepackage{multirow}

\newcommand{\tabincell}[2]{\begin{tabular}{@{}#1@{}}#2\end{tabular}}

\usepackage{newfloat}
\usepackage{listings}
\floatstyle{ruled}
\newfloat{listing}{tb}{lst}{}
\floatname{listing}{Listing}

\title{Improving Zero-shot Multilingual Neural Machine Translation for Low-Resource Languages}

\author {
    Chenyang Li,\textsuperscript{\rm 1}
    Gongxu Luo \textsuperscript{\rm 2}
}
\affiliations {
    \textsuperscript{\rm 1} Beihang University\\
    \textsuperscript{\rm 2} University of Chinese Academy of Sciences\\
    bhselcy@buaa.edu.cn, luogongxu@126.com
}

\date{}

\begin{document}

\maketitle

\begin{abstract}
Although the multilingual Neural Machine Translation(NMT), which extends Google’s multilingual NMT, has ability to perform zero-shot translation and the iterative self-learning algorithm can improve the quality of zero-shot translation, it confronts with two problems: the multilingual NMT model is prone to generate wrong target language when implementing zero-shot translation; the self-learning algorithm, which uses beam search to generate synthetic parallel data, demolishes the diversity of the generated source language and amplifies the impact of the same noise during the iterative learning process. In this paper, we propose the tagged-multilingual NMT model and improve the self-learning algorithm to handle these two problems. Firstly, we extend the Google’s multilingual NMT model and add target tokens to the target languages, which associates the start tag with the target language to ensure that the source language can be translated to the required target language. Secondly, we improve the self-learning algorithm by replacing beam search with random sample to increases the diversity of the generated data and makes it properly cover the true data distribution. Experimental results on IWSLT show that the adjusted tagged-multilingual NMT separately obtains \textbf{9.41} and \textbf{7.85} BLEU scores over the multilingual NMT on 2010 and 2017 Romanian-Italian test sets. Similarly, it obtains \textbf{9.08} and \textbf{7.99} BLEU scores on Italian-Romanian zero-shot translation. Furthermore, the improved self-learning algorithm shows its superiorities over the conventional self-learning algorithm on zero-shot translations.
\end{abstract}

\section{Introduction}

The data driven Neural Machine Translation(NMT), which follows the end-to-end framework, has shown its superiority on high-resource languages in recent years ~\citep{sutskever2014sequence, bahdanau2014neural, wu2016google,vaswani2017attention}. Because of the fact that NMT system is highly depended on extensive high-quality parallel data, which can only be acquired for few language pairs, it is still challenging for low-resource and zero-shot NMT ~\citep{koehn2017six}. Existing approaches for zero-shot NMT include multilingual NMT ~\citep{firat2016zero,ha2016toward,johnson2017google}, interactive multimodal framework ~\citep{kiros2014unifying,nakayama2017zero}, pivot-based NMT ~\citep{wu2007pivot,cheng2016neural,leng2019unsupervised} and teacher-student architecture ~\citep{chen2017teacher}.

We focus on the multilingual NMT system, which is simple and effective. Recent multilingual NMT with a simple approach named target-forcing, which is trained on a mixture of several parallel data and adds a token to the start of the source sentence to determine the target language, is effective for low-resource languages ~\citep{johnson2017google}. Particularly by using this way, the multilingual NMT system has possibility to perform zero-shot translation through sharing the common model. In order to improve the performance of zero-shot languages, \citet{lakew2017improving} proposed the self-learning algorithm, which generates synthetic parallel data by translating existing target data through the multilingual NMT round and round. The whole process is a self-learning cycle of train-infer-train.

However, there still exist some problems about the proposed methods for zero-shot translation. We find that the multilingual NMT system does not accurately translate the source language into the required target language when performing zero-shot translation. In addition, the self-learning algorithm, which uses beam search to choose the highest probability sentence to generate synthetic parallel data, demolishes the diversity of the generated source sentences. Especially in the last few rounds of the iterative process, the model’s effect on zero-shot translation is improved slightly and even declines. We speculate that this is because of the synthetic parallel data that is generated by beam search is almost the same in the last few iteration, which amplifies the effect of harmful noise.

In this paper, we improve the multilingual NMT and the self-learning algorithm to address the two problems. We first extend the Google’s multilingual NMT system and add a token to the start of the target language to indicate that the target language is the required one. This ensures that the source language can be translated into the required target language. Then the multilingual NMT system is trained on the available mixed parallel data until convergence called tagged-multilingual NMT. Next, we improve the self-learning algorithm by replacing beam search with random sample ~\citep{pillemer1988prevalence} to generate synthetic parallel data for zero-shot translation, which can not only increase the diversity of source language of the synthetic parallel data but also can increase fluency in the target language generated by decoder. In order to testify the effectiveness of the improved method, we experiment it on a multilingual-NMT scenario including Italian-English and Romanian-English parallel corpora, assuming the zero-shot translation is Italian-Romanian and Romanian-Italian. For the tagged-multilingual NMT, experimental results show that adding a target tag can not only make the model to accurately translate the source language into the required target language, but also can improve the performance of the multilingual NMT system, especially in zero-shot translations. For the improved self-learning method, the method effectively improve the performance of zero-shot translation and even exceed the single NMT model with 20K parallel data in Romanian-Italian translation.
In summary, our contribution are as follows:
\begin{itemize}
\item	we add a token to the start of the target language to ensure the tagged-multilingual NMT can accurately generate the required target language. It significantly improve the performance of the zero-shot translation and is simultaneously helpful for the low-resource multilingual NMT.
\item	We improve the self-learning method via replacing beam search with random sample to increase the diversity of the generated synthetic parallel data, which makes the generated data more relevant to real language situations.
\item	Experimental result on the multilingual translation shared task published in 2017 International Workshop on Spoken Language Translation(IWSLT) shows the superiorities of our tagged-multilingual NMT model and the improved self-learning method over the previous methods.
\end{itemize}

The reminder of the article is organized as follows. Section 2 summarizes the related work and highlights the differences of our tagged-multilingual NMT model and the improved self-learning algorithm from previous studies. Section 3 briefly describes the NMT model for the multilingual NMT. Section 4 gives details of our proposed tagged-multilingual NMT model and the improved self-learning algorithm. Section 5 introduces the detail of our data sets, experiment settings and baselines. Section 6 reports the experimental results on IWSLT multilingual translation tasks. Finally, we conclude in section 7 with future work.

\section{Related work}
In this section, we first introduce the origin and development of multilingual NMT. Some existing methods for zero-shot translation, which extends multilingual NMT or is based on other architectures, are shown in the next part.
\subsection{Multilingual NMT}
Inspired by the sequence to sequence NMT, \citet{dong2015multi} proposed a one-to-many multi-task NMT to achieve higher translation quality, which has a same source language and different target languages. The same source language can make full use of the source corpora for better representation through a shared encoder. Different target languages use a separate decoder and attention mechanism, which can learn latent similar semantic and structural information across different languages. In a related work, \citet{luong2015multi} used separate encoder and decoder networks for modeling language pairs in a many-to-many setting. Aiming at reducing ambiguity at translation time, \citet{zoph2016multi}  proposed a multi-source NMT with multiple encoders and one attention mechanism, which obtains excellent performance through the novel combination method to encode the common semantic of multiple source languages and the multi-source attention. Follow the Dong and Luong et al’s work, \citet{firat2016multi} proposed a multi-way multilingual NMT, which is a many-to-many translation. It shares the attention mechanism and uses different encoders and decoders across different language pairs. Experimental results show the effectiveness of the shared attention mechanism for low-resource languages.

But, due to the high complexity of the previously mentioned method, Johnson et al and Ha et al attempted to build a multilingual NMT without modifying the network architecture. \citet{ha2016toward} applied a language-specific coding to words of both source and target languages for better representation. In practice, language-specific coding for words and sub-words significantly increased the length of sentences, which causes trouble for sentence representation and attention mechanism. In order to translate into the specific target language, they use target forcing to add a token to the start and the end of the source sentence. Even more concisely, \citet{johnson2017google} just add an artificial token to the start of source sentence to indicate the required target languages. 

\subsection{Zero-shot translation}
Researchers have done fantastic work for zero-shot NMT. An intuitive way is to select a medium as a pivot. \citet{cheng2016neural} proposed a pivot-based method, which use a third language as pivot, for zero-resource NMT. It translate the source language to a pivot language, which is then translated to target language. Similarly, \citet{nakayama2017zero} had shown that multimodal information is also effective as a pivot to zero-resource NMT. However, the pivot method suffers from expensive computational resource and error propagation ~\citep{zhu2013improving}. Base on the exiting problems, \citet{chen2017teacher} proposed a teacher-student architecture for zero-resource NMT by assuming that parallel sentences have close probabilities of generating a sentence in a third language. In ~\citep{chen2018zero}, chen et al proposed a multimodal framework to make full use of monolingual multimodal content to achieve direct modeling of zero-resource source-to-target NMT, which includes captioner and translator two agents. The captioner, which is CNN-RNN architecture, translated image into source sentence. The translator, which is the training target, translated source sentence into the target sentence.
 
The another benefit of multilingual NMT is to have possibility for zero-shot translation. By extending the approach in ~\citep{firat2016multi}, \citet{firat2016zero} extends the one-to-one pivoted based strategy for zero-shot NMT, where the second stage is replaced by the many-to-one strategy. Moreover, the attention mechanism of the many-to-one strategy is fine-tuned by the generated pseudo parallel corpus. Following google’s multilingual NMT, \citet{lakew2017improving} proposed the self-learning algorithm for zero-shot NMT, which is a process of constantly iterating through a train-infer-train mechanism to generate synthetic parallel data for zero-shot translation. By using this way, the quality of the generated parallel data is significantly improved.

Despite the success of the proposed method for zero-shot translation, we extend the method proposed by Lakew et al and found the inadequacies of the multilingual NMT and the self-learning algorithm. Therefore, we improve the multilingual NMT by adding a target token to the start of the target language to indicate the required target language and improve the self-learning method via sampling to increase the diversity of generated parallel data.

\section{Neural machine translation}
Without loss of generality, the multilingual NMT and self-learning method can be applied to any NMT models. We use the transformer model for the multilingual NMT proposed by \citet{vaswani2017attention}, which is by far the most effective end-to-end NMT model. Therefore, in this section, we briefly introduce the overall architecture of the model.

The encoder, which encodes the source sentence $\chi = \left ( \chi _{1},\chi _{2},\cdots ,\chi _{n} \right )$ into a series of context vector $ C = \left ( h _{1},h _{2},\cdots ,h _{n} \right )$, consists of six identical layers. Each layer includes two sub-layers. The first layer is a multi-head attention mechanism, which learns the association between words in a sentence by weighting the sum of all words in the sentence to express a word. The attention mechanism is computed as follows:
\begin{equation}
 Attention\left ( Q,K,V \right )= softmax\left ( \frac{QK^{T}}{\sqrt{d_{k}}} \right )V
\end{equation}
Where $Q$, $K$, $V$ are query, key and value matrix. $QK^{T}$ is used to compute the weights between words. These weights are multiplied by corresponding word embedding and then added to obtain a new representation of the query word in $Q$. Finally, the encoder obtains the context vector $C$ that better presents the source sentence. Compared with RNN and LSTM, self-attention mechanism can better learn long sentence dependencies. The second layer is a full-connected feed forward network. Because of the multi-head attention mechanism map the sentence to different semantic spaces, the six-layer encoder learns the deep semantic relationship of the sentence. 

Similarly, the decoder also consists of six identical layers. The difference is that each layer includes a mask multi-head attention mechanism, attention mechanism and feed forward network. Where the attention mechanism uses the weighted sum of the context vector to represent the target words, which works like word alignment to match the source words with the target words. In the inference process, the target word is generated by maxing
\begin{equation}
    p(y_{t}|y_{<t},C)=softmax(f(y_{<t},C))
\end{equation}
where $y_{<t}$ are the generated words, $f$ represents a series of function operations inside the decoder. Specially, residual connection is applied between the sub-layers to avoid gradient vanishing in both encoder and decoder. At the end of each sub-layer, layer norm is applied to speed up training. 

\section{Tagged-multilingual NMT and the improved self-learning algorithm}
In this section, we describe the tagged-multilingual NMT and the improved self-learning algorithm for zero-shot translation.

We extend the Google’s multilingual NMT and add a token to the start of the target language to indicate the required language. For example, the tagged Roman-English parallel data are as follows:
\begin{flushleft}
Source~(Romanian):\\
$<2en>$ Am zburat cu Air Force Two timp de opt ani.\\
Target~(English):\\
$<en>$ I flew on Air Force Two for eight years.
\end{flushleft}

We add tokens to the source sentences and corresponding target sentences like the example for all parallel data. After this, we train a multilingual NMT model called tagged-multilingual NMT as shown in Figure 1. The decoder starts with a start-tag and then generates the required target tag with the help of attention mechanism when decoding, it ensure that subsequent words are the correct target language, which is more effective for zero-shot translation.

\begin{figure}[htb]

\centering
\includegraphics[width=8.8cm,height=8cm]{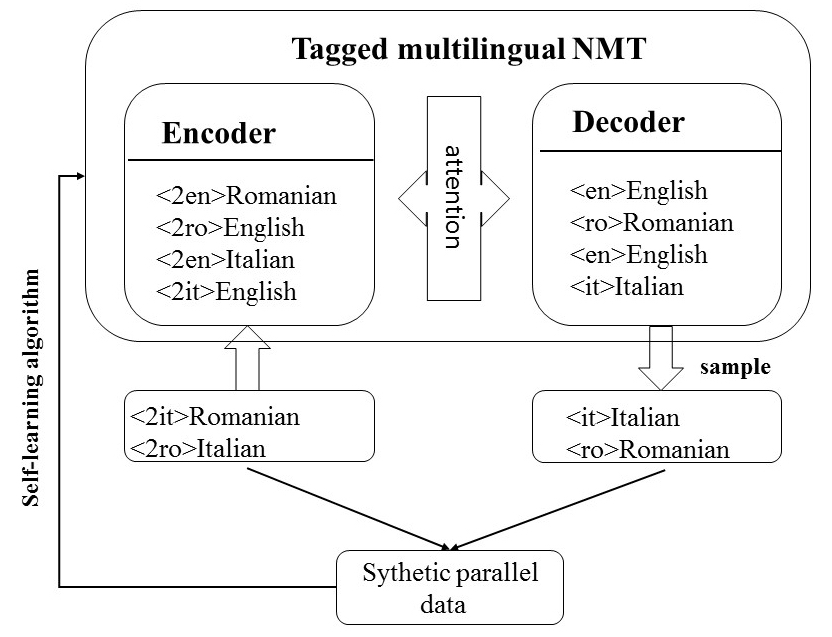}
\caption{The tagged-multilingual NMT model and the improved self-learning algorithm.}
\label{fig1}
\end{figure}

Furthermore, considering the self-learning method with beam search, which choose the sentence with the highest probability, decreases the diversity of the generated synthetic parallel data and amplify the negative effect of noise in the process of continuous iteration, thus we improved the self-learning method via replacing the beam search method with random sample. The improved self-learning method is shown in Algorithm 1.
\begin{algorithm}[ht]
  \caption{Zero-shot translation L1$\leftrightarrow$L2 }
  \label{algorithm1}
  \begin{algorithmic}[1]
   \Require
    mixed parallel data D, source language l1, target language l2
    \State  Tagged-Multilingual NMT $\leftarrow$ Train ($\theta$, D)
    \State  Monolingual L1 $\leftarrow$ Extract form (D, l2)
    \State  Monolingual L2 $\leftarrow$ Extract form (D, l2)
    \For{i=1,N}
    \State  L1* L2* $\leftarrow$ using Tagged-Multilingual NMT to translate L1, L2 to generate the source langue via sampling for zero-shot translation
    \State New mixed parallel data D*=(l1+L1*+L2*, l2+L1+L2)
    \State Update Tagged-Multilingual NTM $\leftarrow$ train($\theta_{1}$, D*) for 3 epoch
    \EndFor
   \Ensure
   Return updated Tagged-Multilingual NMT
  \end{algorithmic}
\end{algorithm}
The improved self-learning methods can be divided into three part. For the first step, we train a tagged-multilingual NMT on the mixed parallel data in line 1. Next, the synthetic parallel data for zero-shot translation is generated by translating the target language into source language through the tagged-multilingual NMT model in line 5, which uses sample to increase the diversity of the synthetic data instead of beam search. The sample during decoding is calculated as follows:
\begin{equation}
    \widetilde{y}_{t}=Sample(f(y_{t}|y_{<t},C))
\end{equation}
The decoder starts with start tag and then generates the next word based on the probability distribution of the word. By using random sample, some low-probability words are generated to create a fluent sentence, which is more fitted to the distribution of real data. Finally, we add the synthetic parallel data to the mixed data to update the tagged-multilingual NMT to get better performance round and round. The iteration is performed a total five times in line 6-7, where the tagged-multilingual NMT is trained on new mixed parallel data for 3 epoch at each time.

\section{Experiment}
In this section, we first introduce the data sets and hyper-parameters for the tagged-multilingual NMT. Next we briefly introduce the baselines that has been mentioned in the prat of related work.
\subsection{Data set and preprocessing}
We consider the scenario that there are Romanian~(Ro) $\leftrightarrow$ English~(En), Italian~(It) $\leftrightarrow$ English~(En) parallel data for the multilingual NMT, assuming that Italian(It) $\leftrightarrow$ Romanian(Ro) are zero-shot translations. The details of the datasets are shown in Table 1. All the parallel data are from the 2017 IWSLT multilingual TED talks machine translation task ~\citep{cettolo2012wit3}. The dev set and test sets are from IWSLT 2010, 2017 evaluation campaigns for development and evaluating the models. Specially, we combine the dev sets of exiting parallel data into a mixed dev set for multilingual NMT.

\begin{table}[htp]
 \centering
  \scriptsize
\caption{All data sets for the tagged-multilingual NMT.}
\begin{tabular}{lcccc}
\hline
\vspace{1mm}\\[-3mm]
Language pair & Train & Dev 2010 & Test 2010 & Test2017 \\
\vspace{1mm}\\[-3mm]
\hline
\vspace{1mm}\\[-3mm]
It-En & 231619 & 929 & 1566 & 1147\\
Ro-En & 220538 & 914 & 1678 & 1129\\
It-Ro & 217551 & 914 & 1643 & 1127\\
\hline
\end{tabular}
\label{table1}
\end{table}

In data preprocessing process, we first use Moses’s word segmentation tool to segment parallel data\footnote{http://www.statmt.org/moses/?n=Moses.Baseline}. Then we segment words into sub-words via Bite Pair Encoding ~\citep{sennrich2016neural} to effectively decrease the number of words that is out-of-vocabulary(OOV). 
\subsection{Experiment settings}
All the experiments are trained based on the transformer models ~\citep{vaswani2017attention}, which is implemented by the Mxnet-based(version 1.4.1) NMT framework sockeye(version 1.18.99) ~\citep{hieber2017sockeye}. We do a lot of works to find suitable hyper-parameters for low-resource languages and the tagged-multilingual NMT model as shown in Table 2 and Table 3. From Table 2, we find that setting BPE merge number to 8000 and drop out to 0.3 gets the best result for It-En translation. However, from Table 3, we can see that embedding dropout is valid for the tagged-multilingual NMT. Especially, it works best on the validation set when embedding dropout is 0.3, and is close to the results when embedding dropout is 0.2. Unfortunately, compared with the embedding dropout of 0.2,it increases the training time by 30$\%$. So, we finally choose embedding dropout is 0.2 and BPE merge number is 12000 for the tagged-multilingual NMT.

\begin{table*}[htp]
 \centering
  \scriptsize
\caption{The hyper-parameters for It-En translation.}
\begin{tabular}{cccccccccccc}
\hline
\vspace{1mm}\\[-3.5mm]
\multicolumn{12}{c}{It-En}\\
\vspace{1mm}\\[-3.5mm]
\hline
\vspace{1mm}\\[-3mm]
BPE merge number & 36000 & 15000 & 10000 & 9000 & 8000 & 8000 & 8000 & 8000 & 7000 & 6000 & 4000\\
Drop out & 0.3 & 0.3 & 0.3 & 0.3 & 0.3 & 0.5 & 0.4 & 0.2 & 0.3 & 0.3 & 0.3\\
Dev set & 25.3 & 27.2 &	28.31 &	27.96 &	28.59 &	26.15 &	28.21 &	27.12 &	28.07 &	28.36 &	27.63\\
\hline
\end{tabular}
\label{table2}
\end{table*}

\begin{table*}[t]
 \centering
  \scriptsize
\caption{The experimental results of the bilingual NMT, the multilingual NMT, the tagged-multilingual NMT and their adjusted model on test 2010 and test 2017.}
\begin{tabular}{cccccccc}
\hline
Direction&\tabincell{c}{Bilingual\\Baseline1} & \tabincell{c}{Multilingual\\Baseline2} & Tagged-multilingual & \tabincell{c}{Adjusted\\ tagged-multilingual} & \tabincell{c}{Adjusted\\ bilingual} & \tabincell{c}{Adjusted\\ multilingual} & \tabincell{c}{Improved\\self-learning\\ algorithm}\\
\hline
\vspace{1mm}\\[-3mm]
\multicolumn{8}{c}{Test2010}\\
\vspace{1mm}\\[-3.5mm]
\hline
\vspace{1mm}\\[-3mm]
It-En&	29.23&	29.71&	30.16(+0.45)&	30.86(+1.15)&	30.51&	30.75 & -\\
En-It&	26.34&	26.23&	26.41(+0.18)&	27.23(+1.00)&	27.00&	26.63 & -\\
Ro-En&	31.53&	31.57&	31.74(+0.17)&	32.94(+1.37)&	32.31&	32.85& -\\
En-Ro&	23.40&	24.03&	24.51(+0.48)&	25.1(+1.07)&	24.01&	24.79&-\\
Ro-It&	19.27&	6.82&	\textbf{10.66(+3.84)}&	\textbf{16.23(+9.41)}&	19.69&	13.71 &19.86\\
It-Ro&	17.60&	6.09&	\textbf{8.78(+2.69)}&	\textbf{15.17(+9.08)}&	18.17&	14.35& 17.96\\
\hline
\vspace{1mm}\\[-3mm]
\multicolumn{8}{c}{Test2017}\\
\vspace{1mm}\\[-3.5mm]
\hline
\vspace{1mm}\\[-3mm]
It-En&	32.20&	32.34&	32.55(+0.21)&	33.31(+0.99)&	33.17&	33.69&-\\
En-It&	28.84&	29.19&	29.16(-0.03)&	30.11(+0.92)&	29.25&	30.03&-\\
Ro-En&	26.11&	27.18&	27.35(+0.17)&	28.15(+0.97)&	27.43&	27.96&-\\
En-Ro&	20.61&	20.88&	21.51(+0.69)&	21.71(0.89)&	20.14&	21.03&-\\
Ro-It&	18.77&	6.82&	\textbf{9.57(+2.75)}&	\textbf{14.67(+7.85)}&	19.51&	12.49& 19.34\\
It-Ro&	17.41&	5.65&	\textbf{7.65(+2)}&	\textbf{13.64(+7.99)}&	17.59&	12.25&16.96\\
\hline
\end{tabular}
\label{table4}
\end{table*}

\begin{table}[htp]
 \centering
  \scriptsize
\caption{The hyper-parameters the tagged-multilingual NMT.}
\begin{tabular}{ccccc}
\hline
\vspace{1mm}\\[-3mm]
\multicolumn{5}{c}{Tagged-multilingual NMT}\\
\vspace{1mm}\\[-3mm]
\hline
\vspace{1mm}\\[-3mm]
BPE merge number & 8000 & 12000 & 12000 & 12000\\
Embedding dropout & 0 & 0 & 0.2 & 0.3\\
Dev set & 19.56 & 20.02 & 20.15 & 20.23\\
\hline
\end{tabular}
\label{table3}
\end{table}

The other hyper-parameters of the transformer are as follows. Considering the high data sparsity of low-resource languages and to prevent over-fitting ~\citep{srivastava2014dropout} of the model, we set the label smoothing ~\citep{szegedy2016rethinking} to 0.1 and set the dropout of 0.3 for multi-head attention, feed-forward network and preprocessing block according to the sennrich et al’s work for low-resource NMT ~\citep{sennrich2019revisiting}. In addition, We use the Adam ~\citep{kingma2014adam} as the optimizer and set the initial learning rate to 0.0003. Particularly, at the beginning of the training, The warmup$-{}$traing$-{}$steps is set to 16000 to warm up the learning rate, which prevents model oscillation caused by random initialization of parameters, batch size is 4096. In the training process, we use early stop as the stop condition of the training model. If the model’s effect on the dev set is no longer improved for 10 times, we contend that the model is optimal. For decoding, a beam search size 10 is applied for the NMT model. Finally, the BLEU score, which is proven evaluation metrics, is applied to verify the effectiveness of the model ~\citep{papineni2002bleu}.
\subsubsection{Baselines}
We refer to our model as the tagged-multilingual NMT and compared it against the following baselines.
\begin{itemize}
    \item Baseline 1 is a bilingual NMT based on transformer.
    \item Baseline 2 is google’s multilingual NMT with self-learning algorithm for zero-shot translation. 
\end{itemize}
The multilingual NMT uses a token at the start of the source language to indicate the required target language. Furthermore, the self-learning algorithm generates synthetic parallel data for zero-shot translation by translating the target language.

\section{Results and analysis}
In this section, we conduct experiment to evaluate the effectiveness of the tagged-multilingual NMT and the improved self-learning method on IWSLT multilingual translation task. We describe the experimental results and analysis them.

\begin{figure*}[t]
    \centering
    \begin{minipage}{6.5cm}
      \includegraphics[width=6.5cm]{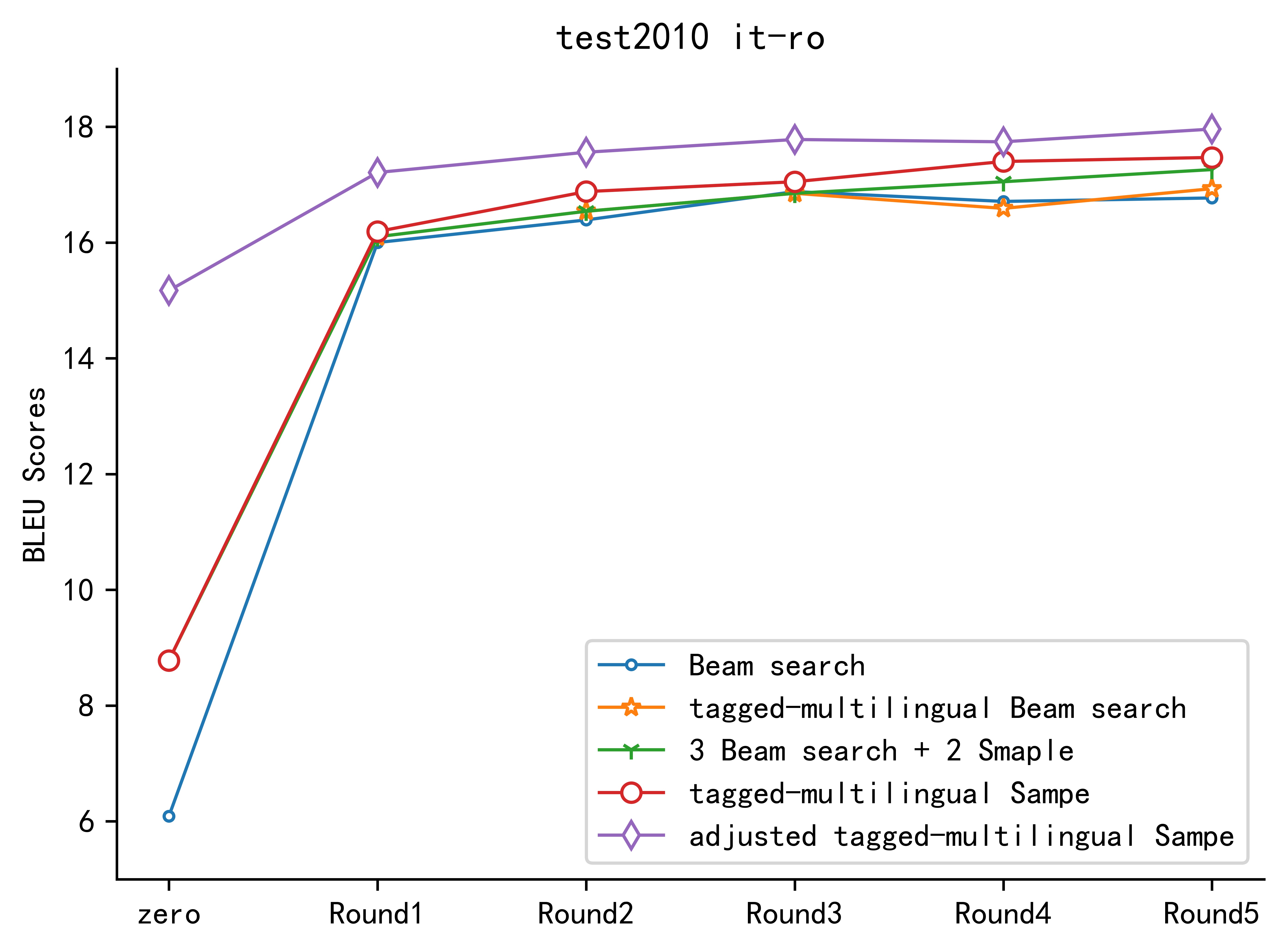}
      \end{minipage}
      \begin{minipage}{6.5cm}
      \includegraphics[width=6.5cm]{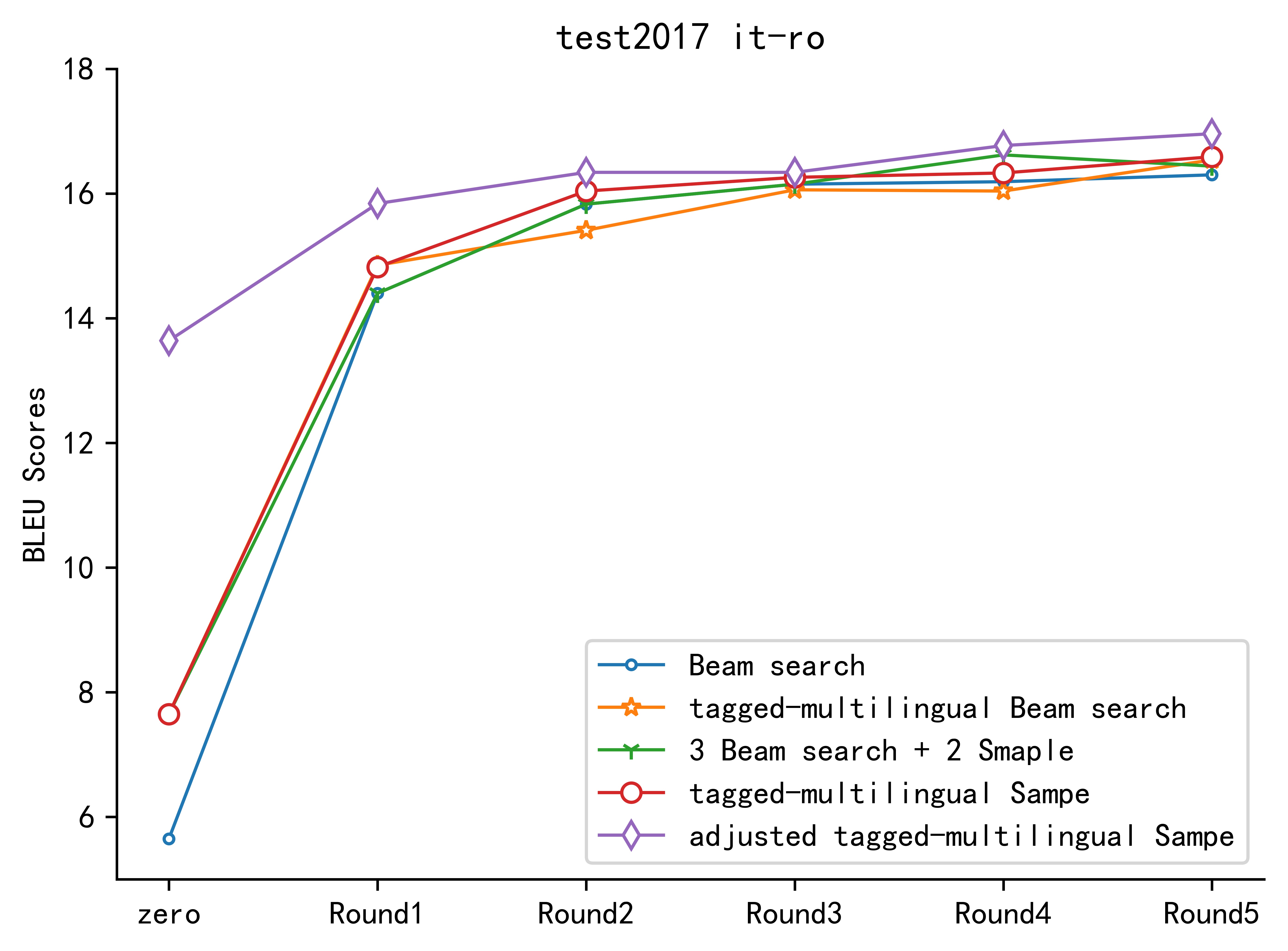}
      \end{minipage}
    \caption{The improved self-learning algorithm for It-Ro zero-shot translation on Test2010 and Test2017.}
    \label{figure2}
\end{figure*}

\begin{figure*}[t]
    \centering
    \begin{minipage}{6.5cm}
      \includegraphics[width=6.5cm]{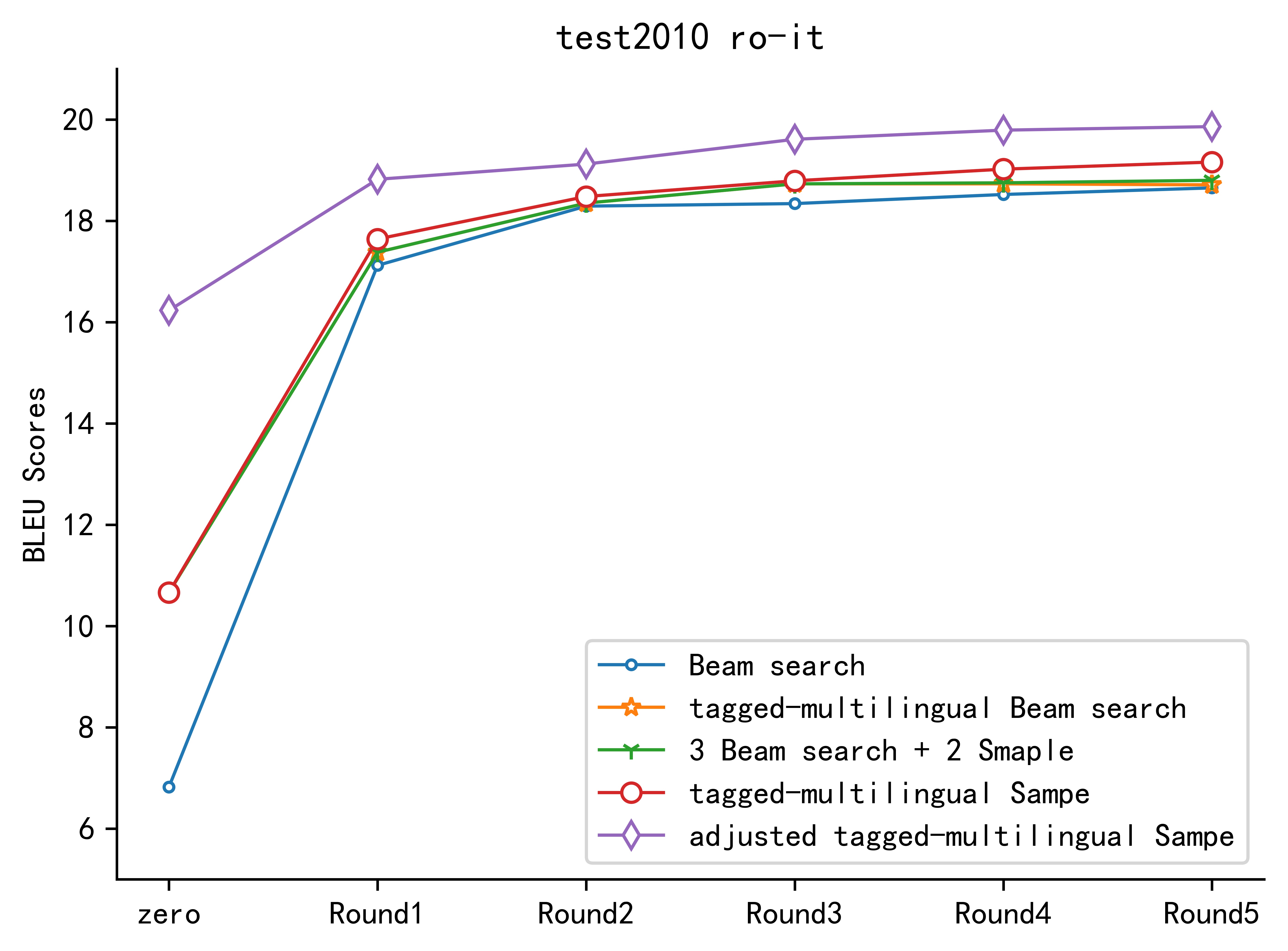}
      \end{minipage}
      \begin{minipage}{6.5cm}
      \includegraphics[width=6.5cm]{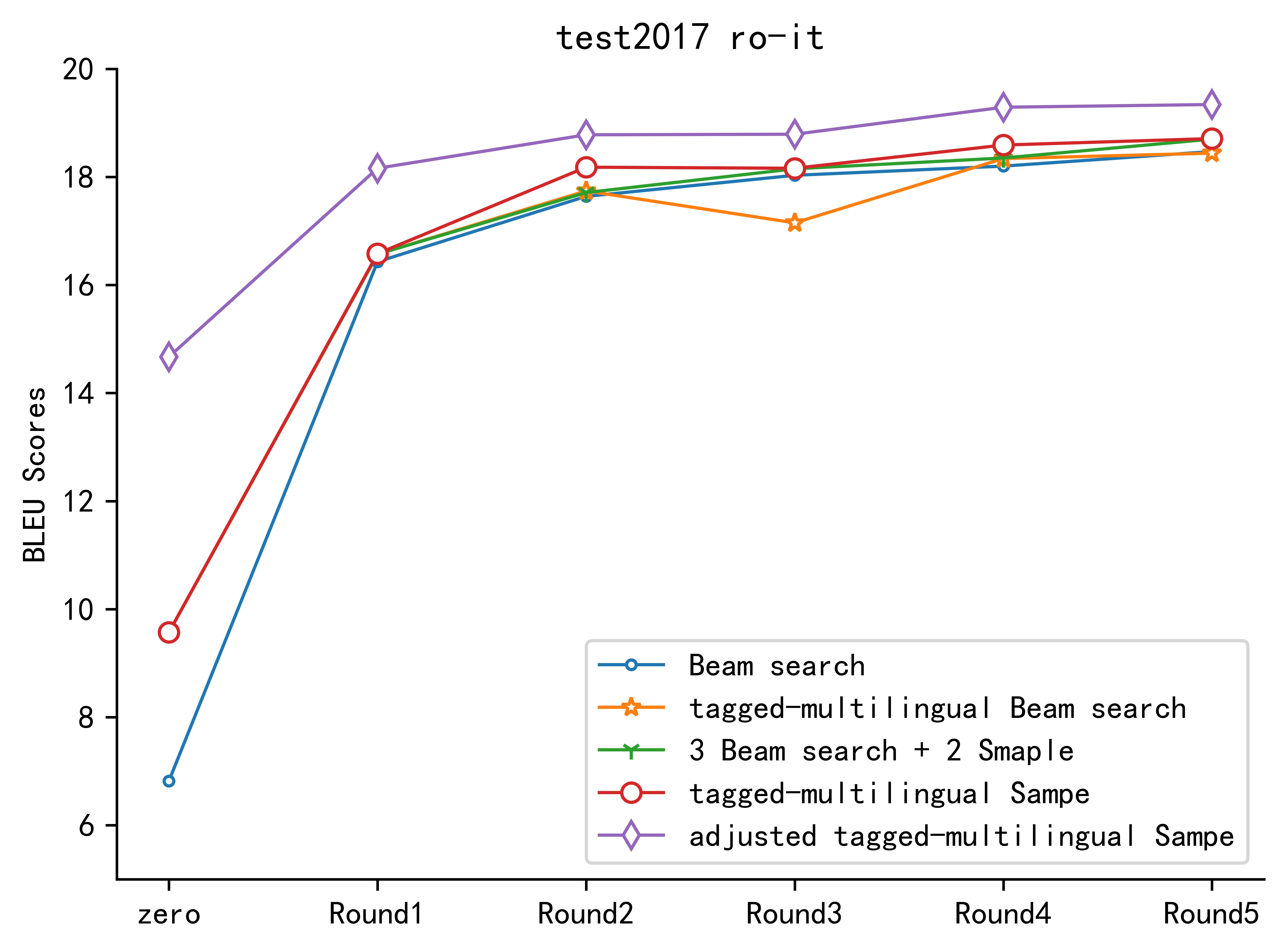}
      \end{minipage}
    \caption{The improved self-learning algorithm for Ro-It zero-shot translation on Test2010 and Test2017}
    \label{figure3}
\end{figure*}

\subsection{Tagged-multilingual NMT}
We train the tagged-multilingual NMT on the mixed parallel data until the performance on the dev set are no longer improved for consecutive ten times. The experimental results of the adjusted tagged-multilingual NMT, the tagged-multilingual NMT, the adjusted multilingual NMT, the multilingual NMT, the adjusted bilingual NMT and the bilingual NMT on test 2010 and 2017are shown in table 4.

From table 4, we can see that the tagged-multilingual NMT achieve the same or better results than the multilingual model. More obviously, the tagged-multilingual NMT improves \textbf{3.84} and \textbf{2.69} BLEU scores respectively for Ro-It and It-Ro zero-shot translations. After tuning the hyperparameters, the tagged-multilingual NMT improves \textbf{9.41} and \textbf{9.08} BLEU scores respectively. Similarly, the results on test 2017 are in same, which improve \textbf{2.75} and \textbf{2} BLEU scores for Ro-It and It-Ro zero-shot translations. Specially the adjusted tagged-multilingual NMT improves \textbf{7.85} and \textbf{7.99} BLEU scores respectively for zero-shot translations. We also find that, for translations with parallel data such as It-En and so on, the multilingual model can translate the source language into required target language. Unfortunately, it works poorly for zero-shot translation, which translates lots of the source language into the target language that has parallel data with the source language. The reasons are that the attention mechanism of the multilingual NMT, which works like word alignment, learns the correspondences between the source language and the target language with corresponding parallel data. More importantly, the source token that can be seen a word of source language dose not have a corresponding target word, which causes the target words and the source token to have low weights in the attention mechanism. Therefore, due to the lack of parallel data, the attention mechanism works poorly for the zero-shot translation. However, the tagged-multilingual model adds corresponding target tokens to the target languages, which corresponds to the source token and associates the start tag with the required target language. In the training process, the self-attention mechanism can learn the relationship between the target token and the target language. The attention mechanism linking source and target languages learns the alignment between the source token and the target token. Therefore, when decoding, the decoder start with the start tag and then generate the target token of the required target language with the help of the attention mechanism. Next, the start tag and the generated target token are presented by self-attention mechanism. Then the required target words will be generated by classifier with the help of attention mechanism. Because the target words are generated by maxing the conditional probability of $p(y|start-tag,target- token, context-vector)$, which bases on the previous words and context vectors, therefore the target token can specify the required target language accurately at the beginning of decoding. Experimental results show that the adding a target token effectively solve the problem of specifying the target language for zero-shot translations.
\subsection{Improved self-learning algorithm}

In order to further improve the performance of zero-shot translation, we improve the self-learning method via replacing the beam search with random sample for the multilingual NMT. We test beam search, sample and a combination of the two methods, where beam search size is 10, sample size is 5. All results on test2010 and test2017 for It-Ro zero-shot translation are shown in Figure 2. Similarly, for Ro-It zero-shot translation, all results of different self-learning algorithms are shown in Figure 3. Besides, the detailed experimental data is shown in appendix. From these Figures, we can see that the effect of the self-learning algorithm with beam search on all test sets increases slowly and even declines since the third round. For this phenomenon, we try to improve it with random sample and the combination of the tow methods. Experimental results as shown in Table 5 and Table 6 in appendix show that both methods are helpful, and the sample is more effective. More obviously, the adjusted self-learning algorithm significantly improves \textbf{1.31} and \textbf{0.66} BLEU scores on test2010 and 2017 It-Ro zero-shot translation, and improves \textbf{1.21} and \textbf{0.87} BLEU scores on test2010 and 2017 Ro-It zero-shot translation. Moreover, compared to adjusted single It-Ro and Ro-It translation with 20K parallel data as shown in table 4, the improved self-learning algorithm for zero-shot translations has very close results with bilingual NMT model and even excess the single model on test2010 Ro-It translation. We contend that beam search chooses the highest probability sentences, which will result in generating same sentences during the iteration. That will demolish the diversity of the generated sentences and amplify the impact of the same noise. However, the synthetic parallel data generated by random sample have sentences that select low probability words with corresponding probability, which is helpful for increasing the diversity of the source language and properly covering the true data distribution. 

\section{Conclusion an future work}
This paper has presented the tagged-multilingual NMT and the improved self-learning algorithm. Unlike the multilingual NMT and the self-learning algorithm, our tagged-multilingual NMT model can correctly translate the source language into the required target language by learning the relationship between the target language and the target token, which is especially obvious in zero-shot translations. The improved self-learning algorithm generates synthetic parallel data closer to the true data distribution through random sample, which increase the diversity of the source language of the synthetic parallel data. Experimental results on IWSLT multilingual NMT tasks and It-Ro bidirectional zero-shot translations show that the tagged-multilingual NMT and the self-learning algorithm achieve significant improvement over a variety of baselines.

In the future work, we are going to explore better approaches to improve the self-learning mechanism to generate high-quality synthetic parallel data.

\bibliography{aaai2022}

\end{document}